\documentclass[11pt]{article}
\usepackage{acl2015}
\usepackage{times}
\usepackage{url}
\usepackage{latexsym}
\usepackage[utf8]{inputenc}
\usepackage{graphicx}
\usepackage{color}
\usepackage[colorinlistoftodos,prependcaption,textsize=tiny]{todonotes}
\usepackage[round]{natbib}
\usepackage[tight,footnotesize]{subfigure}
\usepackage[font=small,margin=0.5cm]{caption}
\usepackage{tabularx}
\usepackage{array}
\usepackage{titlesec}
\usepackage{blindtext}
\usepackage{threeparttable}

\title{Word Embeddings: A Survey}

\author{ Felipe Almeida \qquad Geraldo Xexéo\thanks{\hspace{0.2cm}Geraldo Xexéo is also with the Mathematics Institute (IM-UFRJ), Federal University of Rio de Janeiro} \\Computer and Systems Engineering Program (PESC-COPPE) \\ Federal University of Rio de Janeiro \\ Rio de Janeiro, Brazil \\ {\tt \{falmeida,xexeo\}@cos.ufrj.br} \\}

\begin{document}

\newcolumntype{L}[1]{>{\raggedright\let\newline\\\arraybackslash\hspace{0pt}}p{#1}}
\newcolumntype{C}[1]{>{\centering\let\newline\\\arraybackslash\hspace{0pt}}p{#1}}
\newcolumntype{R}[1]{>{\raggedleft\let\newline\\\arraybackslash\hspace{0pt}}p{#1}}

\maketitle
\begin{abstract}
This work lists and describes the main recent strategies for building fixed-length, dense and distributed representations for words, based on the distributional hypothesis. These representations are now commonly called \textit{word embeddings} and, in addition to encoding surprisingly good syntactic and semantic information, have been proven useful as extra features in many downstream NLP tasks.
\end{abstract}

\section{Introduction}

The task of representing words and documents is part and parcel of most, if not all, Natural Language Processing (NLP) tasks. In general, it has been found to be useful to represent them as vectors, which have an appealing, intuitive interpretation, can be the subject of useful operations (e.g. addition, subtraction, distance measures, etc) and lend themselves well to be used in many Machine Learning (ML) algorithms and strategies.

The Vector Space Model (VSM), generally attributed to Salton (\citeyear{salton_1975}) and stemming from the Information Retrieval (IR) community, is arguably the most successful and influential model to encode words and documents as vectors.

Another very important part of natural language-based solutions is, of course, the study of language models. A language model is a statistical model of language usage. It focuses mainly on predicting the next word given a number of previous words. This is very useful, for instance, in speech recognition software, where one needs to correctly decide what is the word said by the speaker, even when signal quality is poor or there is a lot of background noise.

These two seemingly independent fields have arguably been brought together by recent research on Neural Network Language Models (NNLMs), with \cite{bengio_et_al_2003}) having developed the first\footnote{They claim this idea has been put forward before (\cite{mikkulainen_and_dyer_1991}), but not used at scale.} large-scale language models based on neural nets. 

Their idea was to reframe the problem as an unsupervised learning problem. A key feature of this solution is the way raw words vectors are first projected onto a so-called \textit{embedding layer} before being fed into other layers of the network. Among other reasons, this was imagined to help ease the effect of the curse of dimensionality on language models, and help generalization (\cite{bengio_et_al_2003}).

With time, such \textbf{word embeddings} have emerged as a topic of research in and of themselves, with the realization that they can be used as standalone features in many NLP tasks (\cite{turian_et_al_2010}) and the fact that they encode surprisingly accurate syntactic and semantic word relationships (\cite{linguisticregularities}).

More recently\footnote{Their roots, however, date back at least two decades, with the work of \cite{deerwester_et_al_1990}.}, other ways of creating embeddings have surfaced, which rely not on neural networks and embedding layers but on leveraging word-context matrices to arrive at vector representations for words. Among the most influential models we can cite the GloVe model (\cite{pennington_et_al_2014}).

These two types of model have something in common, namely their reliance on the assumption that words with similar contexts (other words) have the same meaning. This has been called the distributional hypothesis, and has been suggested some time ago by Harris (\citeyear{harris_1954}), among others.

This brings us to the definition of \textit{word embeddings} we will use in this article, as suggested by the literature (for instance, \cite{turian_et_al_2010,blacoe_and_lapata_2012,schnabel_et_al_2015}), according to which word embeddings are \textbf{dense, distributed, fixed-length word vectors, built using word co-occurrence statistics as per the distributional hypothesis}.

Embedding models derived from neural network language models have (\cite{baroni_et_al_2014}) been called \textit{prediction-based} models, since they usually leverage language models, which predict the next word given its context. Other matrix-based models have been called \textit{count-based} models, due to their taking into account global word-context co-occurrence counts to derive word embeddings. \footnote{Note that a link between both types of models has been suggested by \cite{levy_and_goldberg_2014b}.} These are described next.

This survey is structured as follows: in section 2 we describe the origins of statistical language modelling. In section 3 we give an overview of word embeddings, generated both by so-called prediction-based models and by count-based methods. In Section 4 we conclude and in Section 5 we provide some pointers to promising further research topics.

\subsection{Motivation}

To our knowledge, there is no comprehensive survey on word embeddings \footnote{There are, however, systematic studies on the performance of different weighting strategies and distance measures on word-context matrices, authored by Bullinaria and Levy (\citeyear{bullinaria_and_levy_2007,bullinaria_and_levy_2012}).}, let alone one that includes modern developments in this area. Furthermore, we think such a work is useful in the light of the usefulness of word embeddings in a variety of downstream NLP tasks (\cite{turian_et_al_2010}) and strikingly accurate semantic information encoded in such vectors (\cite{linguisticregularities}).

\subsection{Scope}

We chose to include articles/strategies based on a mixture of citation count and reported impact on newer models.

\section{Background: The Vector Space Model and Statistical Language Modelling}

In order to understand the reasons behind the emergence and development of word embeddings, we think two topics are of utmost importance, namely the vector space model and statistical language modelling.

The vector space model is important inasmuch as it underpins a large part of work on NLP; it allows for the use of mature mathematical theory (such as linear algebra and statistics) to support our work. Additionally, vector representations are required for a wide range of machine learning algorithms and methods which are used to help address NLP tasks.

Modern research on word embeddings (particularly prediction-based models) has been, to some extent, borne out of attempts to make language modelling more efficient and more accurate. In fact, word embeddings (\cite{bengio_et_al_2003,bengio_and_senecal_2003, mnih_and_hinton_2007}, to cite a few) have been treated as by-products of language models, and only after some time (arguably after \cite{collobert_and_weston_2008}) has the building of word embeddings been decoupled from the task of language models.

We give brief introductions to these two topics next.

\subsection{The Vector Space Model}

The first problem one encounters when trying to apply analytical methods to text data is probably that of how to represent it in a way that is amenable to operations such as similarity, composition, etc.

One of the earliest approaches to that end was suggested in the field of Information Retrieval (IR), with the work of \cite{salton_1975}. They suggest an encoding procedure whereby each document in a collection is represented by a \textit{t-dimensional} vector, each element representing a distinct term contained in that document. These elements may be binary or real numbers, optionally normalized using a weighting scheme such as TF-IDF, to account for the difference in information provided by each term.

With such a \textit{vector space} in place, one can then proceed onto doing useful work on these vectors, such as calculating the similarity between document vectors (using even simple operations such as the inner-product between them), scoring search results (viewing the search terms as a pseudo document), etc.

Turney and Pantel (\citeyear{turney_and_pantel_2010}) provide a very thorough survey of different ways to leverage the VSM, while explaining the particular applications most suitable for them.

\subsection{Statistical Language Modelling}

Statistical language models are probabilistic models of the distribution of words in a language. For example, they can be used to calculate the likelihood of the next word given the words immediately preceding it (its \textit{context}). One of their earliest uses has been in the field of speech recognition (\cite{bahl_et_al_1983}), to aid in correctly recognizing words and phrases in sound signals that have been subjected to noise and/or faulty channels.

In the realm of textual data, such models are useful in a wide range of NLP tasks, as well as other related tasks, such as information retrieval.

While a full probabilistic model containing the likelihood of every word given all possible word contexts that may arise in a language is clearly intractable, it has been empirically observed that satisfactory results are obtained using a context size as small as 3 words (\cite{goodman_2001}). A simple mathematical formulation of such an \textit{n-gram model} with window size equal to $T$ follows:

\[ P(w_1^T)= \prod_{t=1}^{T} P(w_t | w_1^{t-1}), \]

where $w_t$ is the $t$-th word and $w_i^T$ refers to the sequence of words from $w_i$ to $w_T$, i.e. $(w_i, w_{i+1}, w_{i+2} ... w_T)$. $P(w_t|w_1^{t-1})$ refers to the fraction of times $w_t$ appears after the sequence $w_1^{t-1}$. Actual prediction of the next word given a context is done via maximum likelihood estimation (MLE), over all words in the vocabulary.

Some problems reported with these models have been (\cite{bengio_et_al_2003}) the high dimensionality involved in calculating discrete joint distributions of words with vocabulary sizes in the order of 100,000 words and difficulties related to generalizing the model to word sequences not present in the training set.

Early attempts of mitigating these effects, particularly those related to generalization to unseen phrases, include the use of smoothing, e.g. pretending every new sequence has count one, rather than zero in the training set (this is referred to as \textit{add-one} or \textit{Laplace smoothing}. Also, \textit{backing off} to increasingly shorter contexts when longer contexts aren't available (\cite{katz_1987}). Another strategy which reduces the number of calculations needed and helps with generalization is the clustering of words in so-called \textit{classes} (cf. now famous Brown Clustering \cite{brown_et_al_1992}).

Finally, neural networks (\cite{bengio_et_al_2003, bengio_and_senecal_2003, collobert_and_weston_2008}) and log-linear models (\cite{mnih_and_hinton_2007,efficientestimation, distreprofwords}) have also been used to train language models (giving rise to so-called \textit{neural} language models), delivering better results, as measured by perplexity.

\section{Word Embeddings}

As mentioned before, word embeddings are fixed-length vector representations for words. There are multiple ways to obtain such representations, and this section will explore various different approaches to training word embeddings, detailing and they work and where they differ from each other.

Word embeddings are commonly (\cite{baroni_et_al_2014,pennington_et_al_2014,li_et_al_2015a}) categorized into two types, depending upon the strategies used to induce them. Methods which leverage local data (e.g. a word's context) are called \textbf{prediction-based} models, and are generally reminiscent of neural language models. On the other hand, methods that use global information, generally corpus-wide statistics such as word counts and frequencies are called \textbf{count-based} models. We describe both types next.

\subsection{Prediction-based Models}

The history of the development of prediction-based models for embeddings is deeply linked with that of neural language models (NNLMs), because that is how they were initially produced. As mentioned before, a word's \textit{embedding} is just the projection of the raw word vector into the first layer of such models, the so-called \textit{embedding layer}.

The history of NNLMs, which started with the first large neural language model (\cite{bengio_et_al_2003}), is mostly one of gradual efficiency gains, occasional insights and trade-offs between complex models and simpler models, which can train on more data.

Much though early results (as measured by perplexity) clearly indicated that neural language models were indeed better at modelling language than their previous n-gram-based counterparts, long training times (sometimes upwards of days and weeks) are frequently cited among the major factors that hindered the development of such models.

Not long after the seminal paper by Bengio et al. (\citeyear{bengio_et_al_2003}), many contributions were made towards increasing efficiency and performance of these models. 

\renewcommand{\arraystretch}{1.3}
\begin{table*}[h]
\small
\begin{tabularx}{\textwidth}{|L{15mm}|X|L{15mm}|X|}
\hline
\textbf{Article} & \textbf{Overview of Strategy} & \textbf{Architecture}  & \textbf{Notes} \\ \hline 
Bengio et al. 2003 & Embeddings are derived as a by-product of training a neural network language model. & Neural Net & Commonly referred to as the first neural network language model. \\ \hline
Bengio and Senecal 2003 & Makes improvements on the previous paper, by using a Monte Carlo method to estimate gradients, bypassing the calculation of costly partition functions. & Neural Net & Decreased training times by a factor of 19 with respect to Bengio et al. 2003. \\ \hline
Morin and Bengio 2005 & Full softmax prediction is replaced by a more efficient binary tree approach, where only binary decisions at each node leading to the target word are needed. & Neural Net, Hierarchical Softmax  & Report a speed up with respect to Bengio and Senecal 2003 (over three times as fast during training and 100 times as fast during testing), but at a slightly lower score (perplexity). \\ \hline 
Mnih and Hinton 2007 & Among other models, the log-bilinear model is introduced here. Log-bilinear models are neural networks with a single, linear, hidden layer (\cite{mnih_and_hinton_2008}). & Log-linear Model & First appearance of the log-linear model, which is a simpler model, much faster and slightly outscores the model from Bengio et al. (2003). \\ \hline
Mnih and Hinton 2008 & Authors train the log-bilinear model using hierarchical softmax, as suggested in Morin and Bengio (2005), but the word tree is learned rather than obtained from external sources. & Log-linear Model, Hierarchical Softmax  &  Reports being 200 times as fast as previous log-bilinear models.\\ \hline
Collobert and Weston 2008 & A multi-task neural net is trained using not only unsupervised data but also supervised data such as SRL and POS annotations. The model jointly optimizes all of those tasks, but the target was only to learn embeddings. & Deep Neural Net, Negative Sampling & First time a model was built primarily to output just embeddings. Semi-supervised model (language model + NLP tasks).\\ \hline
Mikolov et al. 2013b & Introduces new two models, namely CBOW and SG. Both are log-linear models, using the two-step training procedure. CBOW predicts the target word given a context, SG predicts each context word given a target word. & Log-linear Model, Hierarchical Softmax & Trained on DistBelief, which is the precursor to TensorFlow (\cite{tensorflow}). \\ \hline
Mikolov et al. 2013c & Improvements to CBOW and SG, including negative sampling instead of hierarchical softmax and subsampling of frequent words. & Log-linear Model, Negative Sampling  & SGNS (skip-gram with negative sampling), the best performing variant of Word2Vec, was introduced here.\\ \hline
Bojanowski et al. 2016 & Embeddings are trained at the n-gram level, in order to help generalization for unseen data, especially for languages where morphology plays an important role. & Log-linear Model, Hierarchical Softmax & Reports better results than SGNS. Embeddings are also reported to be good for composition (into sentence, document embeddings). \\ \hline
\end{tabularx}
\caption{Overview of strategies for building prediction-based models for embeddings.}
\end{table*}

Bengio and Senècal (\citeyear{bengio_and_senecal_2003}) identified that one of the main sources of computational cost was the \textit{partition function} or \textit{normalization factor} required by softmax output layers \footnote{Softmax output layers are used when you train neural networks that need to predict multiple outputs, in this case the probability of each word in the vocabulary being the next word, given the context.}, such as those in neural network language models (NNLMs). Using a concept called \textit{importance sampling} (\cite{doucet_et_al_2001}), they managed to bypass calculation of the costly normalization factor, estimating instead gradients in the neural net using an auxiliary distribution (e.g. old n-gram language models) and sampling random examples from the vocabulary. They report gains of a factor of 19 in training time, with respect to the previous model, with similar scores (as measured by perplexity).

A little bit later, Morin and Bengio\footnote{To our knowledge, this is the first time the term \textit{word embedding} was used in this context.} (\citeyear{morin_and_bengio_2005}) have suggested yet another approach for speeding up training and testing times, using a Hierarchical Softmax layer. They realized that, if one arranged the output words in a hierarchical binary tree structure, one could use, as a proxy for calculating the full distribution for each word, the probability that, at each node leading to the word, the correct path is chosen. Since the height of a binary tree over a set \(V\) of words is \(|V|/\log(|V|)\), this could yield exponential speedup. In practice, gains were less pronounced, but they still managed gains of a factor of 3 for training times and 100 for testing times, w.r.t. the model using importance sampling.

Mnih and Hinton (\citeyear{mnih_and_hinton_2007}) were probably the first authors to suggest the Log-bilinear Model\footnote{These are special cases of \textit{log-linear} models. See Appendix A for more information.} (LBL), which has been very influential in later works as well.  


Another article by Mnih and Hinton (\citeyear{mnih_and_hinton_2008}) can be seen as an extension of the LBL (\cite{mnih_and_hinton_2007}) model, using a slightly modified version of the hierarchical softmax scheme proposed by Morin and Bengio (\citeyear{morin_and_bengio_2005}), yielding a so-called Hierarchical Log-bilinear Model (HLBL). Whereas Morin and Bengio (2005) used a pre-built word tree from WordNet, Mnih and Hinton (\citeyear{mnih_and_hinton_2008}) learned such a tree specifically for the task at hand. In addition to other minor optimizations, they reports large gains over previous LBL models (200 times as fast) and conclude that using purpose-built word trees was key to such results.

Somewhat parallel to the works just mentioned, Collobert and Weston (\citeyear{collobert_and_weston_2008}) approached the problem from a slightly different angle; they were the first to design model with the specific intent of learning embeddings only. In previous models, embeddings were just treated as an interesting by product of the main task (usually language models). In addition to this, they also introduced two improvements worth mentioning: they used words' full contexts (before and after) to predict the centre word \footnote{Previous models focused on building language models, so they just used the left context.}. Perhaps most importantly, they introduced a more clever way of leveraging unlabelled data for producing good embeddings: instead of training a language model (which is not the objective here), they expanded the dataset with \textit{false} or \textit{negative} examples \footnote{I.e. sequences of words with the actual centre word replaced by a random word from the vocabulary.} and simply trained a model that could tell positive (actually occurring) from false examples.\footnote{This has been (\cite{distreprofwords}) called \textit{negative sampling} and speeds up training because one can avoid costly operations such as calculating cross-entropies and softmax terms.}


Here we should mention two specific contributions by Mikolov et al. (\citeyear{mikolov_et_al_2009,mikolov_et_al_2010}), which have been used in later models. In the first work, (\cite{mikolov_et_al_2009}) a two-step method for bootstraping a NNLM was suggested, whereby a first model was trained using a single word as context. Then, the full model (with larger context) was trained, using as initial embeddings those found by the first step.

In (\cite{mikolov_et_al_2010}), the idea of using Recurrent Neural Networks (RNNs) to train language models is first suggested; the argument is that RNNs keep \textit{state} in the hidden layers, helping the model remember arbitrarily long contexts, and one would not need to decide, beforehand, how many words to use as context in either side. 

In \citeyear{mnih_and_teh_2012} Mnih and Teh have suggested further efficiency gains to the training of NNLMs. By leveraging  Noise-contrastive Estimation (NCE). \footnote{Not to be confused with Contrastive Divergence (\cite{hinton_2002}).} NCE (\cite{gutmann_and_hyvarinen_2010}) is a way of estimating probability distributions by means of binary decisions over true/false examples.\footnote{This is somewhat similar to negative sampling, as applied by \cite{collobert_and_weston_2008}. In fact, negative sampling can be seen as a simplified form of NCE, to be used in cases where you just want to train the model (i.e. obtain embeddings), rather than obtain the full probability distribution over the next word (\cite{distreprofwords})}. This has enabled the authors to further reduce training times for NNLMs. In addition to faster training times, they also report better perplexity score w.r.t. previous neural language models. 

It could be said that, in 2013, with Mikolov et al. (\citeyear{linguisticregularities,efficientestimation,distreprofwords}) the NLP community have again (the main other example being \cite{collobert_and_weston_2008}) had its attention drawn to word embeddings as a topic worthy of research in and of itself. These authors analyzed the embeddings obtained with the training of a recurrent neural network model (\cite{mikolov_et_al_2010}) with an eye to finding possible syntactic regularities possibly encoded in the vectors.

\begin{figure}[!htb]
\begin{center}
\includegraphics[width=6cm]{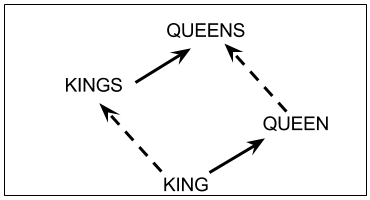}
\caption{Projection of high dimensional word embeddings (obtained with an RNN language model) in 2D: high-level word embeddings encode multiple relationships between words; here shown: singular-plural (dotted line) and male-female (solid line) relationships. Adapted from \cite{linguisticregularities}.}
\vspace{-1.5em}
\label{fig1}
\end{center}
\end{figure}

Perhaps surprisingly, event for the authors themselves, they did find not only syntactic but also semantic regularities in the data. Many common relationships such as male-female, singular-plural, etc actually correspond to arithmetical operations one can perform on word vectors (see Figure \ref{fig1} for an example).

A little later, in \citeyear{efficientestimation} and \citeyear{distreprofwords}, Mikolov et al. have introduced two models for learning embeddings, namely the continuous bag-of-words (\textbf{CBOW}) and skip-gram (\textbf{SG}) models. Both of these models are log-linear models (as seen in previous works) and use the two-step procedure (\cite{mikolov_et_al_2009}) for training. The main difference between CBOW and SG lies in the loss function used to update the model; while CBOW trains a model that aims to predict the centre word based upon its context, in SG the roles are reversed, and the centre word is, instead, used to predict each word appearing in its context.

\begin{table*}[h]
\small
\begin{tabularx}{\textwidth}{|L{15mm}|X|L{60mm}|}
\hline
\textbf{Article} & \textbf{Overview of Strategy} &  \textbf{Notes} \\ \hline 
Deerwester et al. 1990 & LSA is introduced. Singular value decomposition (SVD) is applied on a term-document matrix. &  Used mostly for IR, but can be used to build word embeddings. \\ \hline
Lund and Burgess 1996 & The HAL method is introduced. Scan the whole corpus one word at a time, with a context window around the word to collect weighted word-word co-occurrence counts, building a word-word co-occurrence matrix. & Reported an optimal context size of 8. \\ \hline
Rohde et al. 2006 & Authors introduce the COALS method, which is an improved version of HAL, using normalization procedures to stop very common terms from overly affecting co-occurrence counts.  & Optimal variant used SVD factorization. Reports gains over HAL (\cite{lund_and_burgess_1996}), LSA (\cite{deerwester_et_al_1990}) and other methods. \\ \hline
Dhillon et al. 2011 & LR-MVL is introduced. Uses CCA (Canonical Correlation Analysis) between left and right contexts to induce word embeddings. & Reports gains over C\&W embeddings (\cite{collobert_and_weston_2008}), HLBL (\cite{mnih_and_hinton_2008}) and other methods, over many NLP tasks. \\ \hline
Lebret and Collobert 2013 & Applied a modified version of Principal Component Analysis (called Hellinger PCA) to the word-context matrix. & Embeddings can be \textit{tuned} before being used in actual NLP tasks. Also reports gains over C\&W embeddings, HLBL and other methods, over many NLP tasks. \\ \hline
Pennington et al. 2014 & Introduced \textit{GloVe}, a log-linear model trained to encode semantic relationships between words as vector offsets in the learned vector space, using the insight that co-occurrence ratios, rather than raw counts, are the actual conveyors of word meaning. & Reports gains over all previous count-based models and also SGNS (\cite{distreprofwords}), in multiple NLP tasks. \\ \hline
\end{tabularx}
\caption{Overview of strategies for building count-based models for embeddings.}
\end{table*}

The first versions of CBOW and SG (\cite{efficientestimation}) use hierarchical softmax layers, while the variants\footnote{These have been published under the popular \textit{Word2Vec} toolkit (https://code.google.com/archive/p/word2vec/).} suggested in \cite{distreprofwords} use negative sampling instead. Furthermore, the variants introduced subsampling of frequent words, to reduce the amount of noise due to overly frequent words and accelerate training. These variants were shown to perform better, with faster training times.

Among the most recent contributions to prediction-based models for word embeddings one can cite the two articles (\cite{bojanowski_et_al_2016} and \cite{joulin_et_al_2016}) usually cited as the sources of the \textit{FastText}\footnote{https://research.fb.com/projects/fasttext/} toolkit, made available by Facebook, Inc. They have suggested an improvement over the skip-gram model from \cite{distreprofwords}, whereby one learns not word embeddings, but \textit{n-gram} embeddings (which can be composed to form words). The rationale behind this decision lies in the fact that languages that rely heavily on morphology and compositional word-building (such as Turkish, Finnish and other highly inflexional languages) have some information encoded in the word parts themselves, which can be used to help generalize to unseen words. They report better results w.r.t. SGNS (skip-gram variant with negative sampling) (\cite{distreprofwords}), particularly in languages such as German, French and Spanish.

A structured comparison of prediction-based models for building word embeddings can be seen on Table 1.

\subsection{Count-based Models}

As mentioned before, count-based models are another way of producing word embeddings, not by training algorithms that predict the next word given its context (as is the case in language modelling) but by leveraging word-context co-occurrence counts globally in a corpus. These are very often represented (\cite{turney_and_pantel_2010}) as word-context matrices.

The earliest relevant example of leveraging word-context matrices to produce word embeddings is, of course, \textit{Latent Semantic Analysis} (LSA) (\cite{deerwester_et_al_1990}) where SVD is applied to a term-document \footnote{Term-document matrices are a subset of word-context matrices \cite{turney_and_pantel_2010}.} matrix. This solution was initially envisioned to help with information retrieval. While one is probably more interested in document vectors in IR, it's also possible to obtain word vectors this way; one just needs to look at the rows (rather than columns) of the factorized matrix.

A little later, \cite{lund_and_burgess_1996} have introduced the \textit{Hyperspace Analogue to Language} (HAL). Their strategy can be described as follows: for each word in the vocabulary, analyze all \textit{contexts} it appears in and calculate the co-occurrence count between the target word and each context word, inversely proportional to the distance from the context word to the target word. The authors report good results (as measured by analogy tasks), with an optimal context window size of 8.

The original HAL model did not apply any normalization to word co-occurrence counts found. Therefore, very common words like \textit{the} contribute disproportionately to all words that co-occur with them. \cite{rohde_et_al_2006} have found this to be a problem, and introduced the \textit{COALS} method, introducing normalization strategies to factor out such frequency differences in words. Instead of using raw counts, they suggest it's better to consider the \textit{conditional} co-occurrence, i.e. how much more more likely a word $a$ is to co-occur with word $b$ than it is to co-occur with a random word from the vocabulary. They report better results than previous methods, using the SVD-factorized variant\footnote{I.e., factorizing the co-occurrence matrix in order to reduce dimensions and improve results.}.

A somewhat different alternative was proposed by \cite{dhillon_et_al_2011}, in which they introduce the \textit{Low Rank Multi-View Learning} (LR-MVL) method. In short, it's an iterative algorithm where embeddings are derived by leveraging Canonical Correlation Analysis (CCA) (\cite{hotelling_1935}) between the left and right contexts of a given word. One interesting feature of this model is that when embeddings are used for downstream NLP tasks, they are concatenated with embeddings for their context words too, yielding better results. Authors report gains over other matrix factorization methods, as well as neural embeddings, over many NLP tasks.

\cite{lebret_and_collobert_2013} have also contributed to count-based models by suggesting that a Hellinger PCA\footnote{This amounts to minimizing the distance between principal components and actual data, but using the Hellinger distance instead of the more common Euclidean distance.} transformation be applied to the word-context matrix instead. Results are reported to be better than previous count-based models such as LR-MVL and neural embeddings, such as those by \cite{collobert_and_weston_2008} and HLBL \cite{mnih_and_hinton_2008}.

The last model we will cover in this section is the well-known \textit{GloVe}\footnote{https://nlp.stanford.edu/projects/glove/} by \cite{pennington_et_al_2014}. This model starts at the insight that \textit{ratios} of co-occurrences, rather than raw counts, encode actual semantic information about pair of words. This relationship is used to derive a suitable loss function for a log-linear model, which is then trained to maximize the similarity of every word pair, as measured by the ratios of co-occurrences mentioned earlier. Authors report better results than other count-based models, as well as prediction based models such as SGNS (\cite{distreprofwords}), in tasks such as word analogy and NER (named entity recognition).

A structured comparison of count-based models for building word embeddings can be seen on Table 2.



\section{Conclusion}

\textit{Word embeddings} have been found to be very useful for many NLP tasks, including but not limited to Chunking (\cite{turian_et_al_2010}), Question Answering (\cite{tellex_et_al_2003}), Parsing and Sentiment Analysis (\cite{socher_et_al_2011}).

We have here outlined some of the main works and approaches used so far to derive these embeddings, both using \textit{prediction-based} models, which model the probability of the next word given a sequence of words (as is the case with language models) and \textit{count-based} models, which leverage global co-occurrence statistics in word-context matrices.

Many of the suggested advances seen in the literature have been incorporated in widely used toolkits, such as \textit{Word2Vec}, \textit{gensim}\footnote{https://radimrehurek.com/gensim/}, \textit{FastText}, and \textit{GloVe}, resulting in ever more accurate and faster word embeddings, ready to be used in NLP tasks.

\section{Further Work}

Research on the topic of word representations (and word embeddings in particular) is still active; among the most promising research directions we consider:

\subsection{Adapting embeddings for task-specific work}

Works such as \cite{maas_et_al_2011}, \cite{labutov_and_lipson_2013} and \cite{lebret_and_collobert_2013} have highlighted improved results for NLP tasks when embeddings are \textit{tuned} for specific tasks. 

\subsection{The link between prediction-based and count-based models}

For example, \cite{levy_and_goldberg_2014b} have suggested that the SGNS model (\cite{distreprofwords}) actually is equivalent to using a slightly modified word-context matrix, weighted using PMI (pointwise mutual information) statistics. Insight on what links the two models may yield more advances in both areas. 

\subsection{Composing word embeddings for higher-level entities}

While research on how to compose word vectors to represent higher-level entities such as sentences and documents is not altogether new (generally under the name of \textit{distributional compositionality}), recent works have adapted solutions specifically for neural word embeddings: we can cite here \textit{Paragraph2Vec} (\cite{le_and_mikolov_2014}), \textit{Skip-Thought Vectors} by \cite{kiros_et_al_2015} and also \textit{FastText} itself (\cite{joulin_et_al_2016} and \cite{bojanowski_et_al_2016}).

\appendix

\section{Log-linear Models and Neural Embeddings}

Log-linear models are probabilistic devices which can be used to model conditional probabilities, much like those between word contexts and target words, these being the fundamental parts of language models. 

Log linear models subscribe to the following template (\cite{collins_log_linear_tutorial}) for each output unit:

$$P(y \ | \ x;v) = \frac{ exp(v \ \cdot \ f(x,y)) }{ \sum_{y' \in Y}{{ exp(v \ \cdot \ {f(x,y')}}}) }$$

As applied to the language modelling task, with neural embeddings: $y$ represents the label, i.e., a target word. $x$ represents a word context, i.e. the words before or around the target word we want to predict. $v$ is a learned parameter, i.e. a single row vector in the shared weight matrix.

It's possible to view the formulation above as a neural network with a single, \textit{linear} \footnote{The exponential function is only used to force the values to be positive, and can be removed without loss of generality.} hidden layer, linked to a softmax output layer. Furthermore, akin to any neural-network model, this also can be trained with gradient-based methods, be extended to include regularization terms, and so on.

\bibliographystyle{plainnatsimple}
\bibliography{ref}

\begin{thebibliography}{48}
\providecommand{\natexlab}[1]{#1}
\providecommand{\url}[1]{\texttt{#1}}
\expandafter\ifx\csname urlstyle\endcsname\relax
  \providecommand{\doi}[1]{doi: #1}\else
  \providecommand{\doi}{doi: \begingroup \urlstyle{rm}\Url}\fi

\bibitem[Abadi et~al.(2015)Abadi, Agarwal, Barham, et~al.]{tensorflow}
Mart\'{\i}n Abadi, Ashish Agarwal, Paul Barham, et~al., 2015.
\newblock {TensorFlow}: Large-scale machine learning on heterogeneous systems.
\newblock Available at \url{http://tensorflow.org/}.
\newblock Software available from tensorflow.org.

\bibitem[Bahl et~al.(1983)Bahl, Jelinek, and Mercer]{bahl_et_al_1983}
L.~R. Bahl, F.~Jelinek, and R.~L. Mercer, March 1983.
\newblock A maximum likelihood approach to continuous speech recognition.

\bibitem[Baroni et~al.(2014)Baroni, Dinu, and Kruszewski]{baroni_et_al_2014}
Marco Baroni, Georgiana Dinu, and Germ\'{a}n Kruszewski.
\newblock June 2014.
\newblock Don't count, predict! a systematic comparison of context-counting vs.
  context-predicting semantic vectors.
\newblock In \emph{Proceedings of the 52nd Annual Meeting of the Association
  for Computational Linguistics (Volume 1: Long Papers)}. Association for
  Computational Linguistics.

\bibitem[Bengio and Senécal(2003)]{bengio_and_senecal_2003}
Yoshua Bengio and Jean-Sébastien Senécal, 2003.
\newblock Quick training of probabilistic neural nets by importance sampling.

\bibitem[Bengio et~al.(2003)Bengio, Ducharme, Vincent, and
  Janvin]{bengio_et_al_2003}
Yoshua Bengio, Jean Ducharme, Pascal Vincent, and Christian Janvin, March 2003.
\newblock A neural probabilistic language model.
\newblock Available at \url{http://dl.acm.org/citation.cfm?id=944919.944966}.

\bibitem[Blacoe and Lapata(2012)]{blacoe_and_lapata_2012}
William Blacoe and Mirella Lapata.
\newblock July 2012.
\newblock A comparison of vector-based representations for semantic
  composition.
\newblock In \emph{Proceedings of the 2012 Joint Conference on Empirical
  Methods in Natural Language Processing and Computational Natural Language
  Learning}. Association for Computational Linguistics.

\bibitem[Bojanowski et~al.(2016)Bojanowski, Grave, Joulin, and
  Mikolov]{bojanowski_et_al_2016}
Piotr Bojanowski, Edouard Grave, Armand Joulin, and Tomas Mikolov, 2016.
\newblock Enriching word vectors with subword information.
\newblock Available at \url{http://arxiv.org/abs/1607.04606}.

\bibitem[Brown et~al.(1992)Brown, deSouza, Mercer, Pietra, and
  Lai]{brown_et_al_1992}
Peter~F. Brown, Peter~V. deSouza, Robert~L. Mercer, Vincent J.~Della Pietra,
  and Jenifer~C. Lai, 1992.
\newblock Class-based n-gram models of natural language.

\bibitem[Bullinaria and Levy(2007)]{bullinaria_and_levy_2007}
John~A. Bullinaria and Joseph~P. Levy, 2007.
\newblock Extracting semantic representations from word co-occurrence
  statistics: A computational study.

\bibitem[Bullinaria and Levy(2012)]{bullinaria_and_levy_2012}
John~A. Bullinaria and Joseph~P. Levy, 2012.
\newblock Extracting semantic representations from word co-occurrence
  statistics: stop-lists, stemming, and svd.
\newblock Available at \url{http://dx.doi.org/10.3758/s13428-011-0183-8}.

\bibitem[Collins()]{collins_log_linear_tutorial}
Michael Collins.
\newblock Log-linear models.
\newblock Available at
  \url{http://www.cs.columbia.edu/~mcollins/loglinear.pdf}.

\bibitem[Collobert and Weston(2008)]{collobert_and_weston_2008}
R.~Collobert and J.~Weston.
\newblock 2008.
\newblock A unified architecture for natural language processing: Deep neural
  networks with multitask learning.
\newblock In \emph{International Conference on Machine Learning, {ICML}}.

\bibitem[Deerwester et~al.(1990)Deerwester, Dumais, Furnas, Landauer, and
  Harshman]{deerwester_et_al_1990}
Scott Deerwester, Susan~T. Dumais, George~W. Furnas, Thomas~K. Landauer, and
  Richard Harshman, 1990.
\newblock Indexing by latent semantic analysis.

\bibitem[Dhillon et~al.(2011)Dhillon, Foster, and Ungar]{dhillon_et_al_2011}
Paramveer Dhillon, Dean~P Foster, and Lyle~H. Ungar.
\newblock 2011.
\newblock Multi-view learning of word embeddings via cca.
\newblock In J.~Shawe-Taylor, R.~S. Zemel, P.~L. Bartlett, F.~Pereira, and
  K.~Q. Weinberger, editors, \emph{Advances in Neural Information Processing
  Systems 24}. Curran Associates, Inc.

\bibitem[Doucet(2001)]{doucet_et_al_2001}
Arnaud Doucet.
\newblock 2001.
\newblock \emph{Sequential Monte Carlo methods in practice}.
\newblock Springer, New York.

\bibitem[Goodman(2001)]{goodman_2001}
Joshua Goodman, 2001.
\newblock Classes for fast maximum entropy training.
\newblock Available at \url{http://arxiv.org/abs/cs.CL/0108006}.

\bibitem[Gutmann and Hyvärinen(2010)]{gutmann_and_hyvarinen_2010}
Michael Gutmann and Aapo Hyvärinen, 2010.
\newblock Noise-contrastive estimation: A new estimation principle for
  unnormalized statistical models.

\bibitem[Harris(1954)]{harris_1954}
Zellig~S. Harris, 1954.
\newblock Distributional structure.
\newblock Available at \url{http://dx.doi.org/10.1080/00437956.1954.11659520}.

\bibitem[Hinton(2002)]{hinton_2002}
Geoffrey~E. Hinton, August 2002.
\newblock Training products of experts by minimizing contrastive divergence.
\newblock Available at \url{http://dx.doi.org/10.1162/089976602760128018}.

\bibitem[Hotelling(1935)]{hotelling_1935}
Harold Hotelling, 1935.
\newblock Canonical correlation analysis (cca).

\bibitem[Joulin et~al.(2016)Joulin, Grave, Bojanowski, and
  Mikolov]{joulin_et_al_2016}
Armand Joulin, Edouard Grave, Piotr Bojanowski, and Tomas Mikolov, 2016.
\newblock Bag of tricks for efficient text classification.
\newblock Available at \url{http://arxiv.org/abs/1607.01759}.

\bibitem[Katz(1987)]{katz_1987}
Slava~M. Katz.
\newblock 1987.
\newblock Estimation of probabilities from sparse data for the language model
  component of a speech recognizer.
\newblock In \emph{IEEE Transactions on Acoustics, Speech and Signal
  Processing}.

\bibitem[Kiros et~al.(2015)Kiros, Zhu, Salakhutdinov, Zemel, Torralba, Urtasun,
  and Fidler]{kiros_et_al_2015}
Ryan Kiros, Yukun Zhu, Ruslan Salakhutdinov, Richard~S. Zemel, Antonio
  Torralba, Raquel Urtasun, and Sanja Fidler, 2015.
\newblock Skip-thought vectors.
\newblock Available at \url{http://arxiv.org/abs/1506.06726}.

\bibitem[Labutov and Lipson(2013)]{labutov_and_lipson_2013}
Igor Labutov and Hod Lipson, 2013.
\newblock Re-embedding words.

\bibitem[Le and Mikolov(2014)]{le_and_mikolov_2014}
Quoc~V. Le and Tomas Mikolov, 2014.
\newblock Distributed representations of sentences and documents.
\newblock Available at \url{http://arxiv.org/abs/1405.4053}.

\bibitem[Lebret and Collobert(2013)]{lebret_and_collobert_2013}
R{\'{e}}mi Lebret and Ronan Collobert, 2013.
\newblock Word emdeddings through hellinger {PCA}.
\newblock Available at \url{http://arxiv.org/abs/1312.5542}.

\bibitem[Levy and Goldberg(2014)]{levy_and_goldberg_2014b}
Omer Levy and Yoav Goldberg.
\newblock 2014.
\newblock Neural word embedding as implicit matrix factorization.
\newblock In \emph{Advances in Neural Information Processing Systems 27: Annual
  Conference on Neural Information Processing Systems 2014, December 8-13 2014,
  Montreal, Quebec, Canada}.

\bibitem[Li et~al.(2015)Li, Zhu, and Miao]{li_et_al_2015a}
Shaohua Li, Jun Zhu, and Chunyan Miao, 2015.
\newblock A generative word embedding model and its low rank positive
  semidefinite solution.
\newblock Available at \url{http://arxiv.org/abs/1508.03826}.

\bibitem[Lund and Burgess(1996)]{lund_and_burgess_1996}
Kevin Lund and Curt Burgess, 1996.
\newblock Producing high-dimensional semantic spaces from lexical
  co-occurrence.
\newblock Available at \url{http://dx.doi.org/10.3758/BF03204766}.

\bibitem[Maas et~al.(2011)Maas, Daly, Pham, Huang, Ng, and
  Potts]{maas_et_al_2011}
Andrew~L. Maas, Raymond~E. Daly, Peter~T. Pham, Dan Huang, Andrew~Y. Ng, and
  Christopher Potts.
\newblock 2011.
\newblock Learning word vectors for sentiment analysis.
\newblock In \emph{Proceedings of the 49th Annual Meeting of the Association
  for Computational Linguistics: Human Language Technologies - Volume 1}, HLT
  '11. Association for Computational Linguistics.

\bibitem[Miikkulainen and Dyer(1991)]{mikkulainen_and_dyer_1991}
Risto Miikkulainen and Michael~G. Dyer, 7 1991.
\newblock Natural language processing with modular pdp networks and distributed
  lexicon.
\newblock Available at
  \url{http:https://dx.doi.org/10.1207/s15516709cog1503_2}.

\bibitem[Mikolov et~al.(2009)Mikolov, Kopecky, Burget, Glembek, and
  Cernocky]{mikolov_et_al_2009}
Tomas Mikolov, Jiri Kopecky, Lukas Burget, Ondrej Glembek, and Jan Cernocky.
\newblock 2009.
\newblock Neural network based language models for highly inflective languages.
\newblock In \emph{Proceedings of the 2009 IEEE International Conference on
  Acoustics, Speech and Signal Processing}, ICASSP '09. IEEE Computer Society.

\bibitem[Mikolov et~al.(2010)Mikolov, Karafi{\'{a}}t, Burget, Cernock{\'{y}},
  and Khudanpur]{mikolov_et_al_2010}
Tomas Mikolov, Martin Karafi{\'{a}}t, Luk{\'{a}}s Burget, Jan Cernock{\'{y}},
  and Sanjeev Khudanpur.
\newblock 2010.
\newblock Recurrent neural network based language model.
\newblock In \emph{{INTERSPEECH} 2010, 11th Annual Conference of the
  International Speech Communication Association, Makuhari, Chiba, Japan,
  September 26-30, 2010}.

\bibitem[Mikolov et~al.(2013a)Mikolov, Yih, and Zweig]{linguisticregularities}
Tomas Mikolov, Wen{-}tau Yih, and Geoffrey Zweig.
\newblock 2013a.
\newblock Linguistic regularities in continuous space word representations.
\newblock In \emph{Human Language Technologies: Conference of the North
  American Chapter of the Association of Computational Linguistics,
  Proceedings, June 9-14, 2013, Westin Peachtree Plaza Hotel, Atlanta, Georgia,
  {USA}}.

\bibitem[Mikolov et~al.(2013b)Mikolov, Chen, Corrado, and
  Dean]{efficientestimation}
Tomas Mikolov, Kai Chen, Greg Corrado, and Jeffrey Dean, 2013b.
\newblock Efficient estimation of word representations in vector space.
\newblock Available at \url{http://arxiv.org/abs/1301.3781}.

\bibitem[Mikolov et~al.(2013c)Mikolov, Sutskever, Chen, Corrado, and
  Dean]{distreprofwords}
Tomas Mikolov, Ilya Sutskever, Kai Chen, Gregory~S. Corrado, and Jeffrey Dean.
\newblock 2013c.
\newblock Distributed representations of words and phrases and their
  compositionality.
\newblock In \emph{Advances in Neural Information Processing Systems 26: 27th
  Annual Conference on Neural Information Processing Systems 2013. Proceedings
  of a meeting held December 5-8, 2013, Lake Tahoe, Nevada, United States.}

\bibitem[Mnih and Hinton(2007)]{mnih_and_hinton_2007}
Andriy Mnih and Geoffrey Hinton.
\newblock 2007.
\newblock Three new graphical models for statistical language modelling.
\newblock In \emph{ICML '07: Proceedings of the 24th international conference
  on Machine learning}. ACM.

\bibitem[Mnih and Hinton(2008)]{mnih_and_hinton_2008}
Andriy Mnih and Geoffrey Hinton.
\newblock 2008.
\newblock A scalable hierarchical distributed language model.
\newblock In \emph{In NIPS}.

\bibitem[Mnih and Teh(2012)]{mnih_and_teh_2012}
Andriy Mnih and Yee~Whye Teh.
\newblock 2012.
\newblock A fast and simple algorithm for training neural probabilistic
  language models.
\newblock In \emph{In Proceedings of the International Conference on Machine
  Learning}.

\bibitem[Morin and Bengio(2005)]{morin_and_bengio_2005}
Frederic Morin and Yoshua Bengio.
\newblock 2005.
\newblock Hierarchical probabilistic neural network language model.
\newblock In Robert~G. Cowell and Zoubin Ghahramani, editors, \emph{Proceedings
  of the Tenth International Workshop on Artificial Intelligence and
  Statistics}. Society for Artificial Intelligence and Statistics.

\bibitem[Pennington et~al.(2014)Pennington, Socher, and
  Manning]{pennington_et_al_2014}
Jeffrey Pennington, Richard Socher, and Christopher Manning.
\newblock October 2014.
\newblock Glove: Global vectors for word representation.
\newblock In \emph{Proceedings of the 2014 Conference on Empirical Methods in
  Natural Language Processing (EMNLP)}. Association for Computational
  Linguistics.

\bibitem[Rohde et~al.(2006)Rohde, Gonnerman, and Plaut]{rohde_et_al_2006}
Douglas L.~T. Rohde, Laura~M. Gonnerman, and David~C. Plaut, 2006.
\newblock An improved model of semantic similarity based on lexical
  co-occurence.

\bibitem[Salton et~al.(1975)Salton, Wong, and Yang]{salton_1975}
G.~Salton, A.~Wong, and C.~S. Yang, November 1975.
\newblock A vector space model for automatic indexing.
\newblock Available at \url{http://doi.acm.org/10.1145/361219.361220}.

\bibitem[Schnabel et~al.(2015)Schnabel, Labutov, Mimno, and
  Joachims]{schnabel_et_al_2015}
Tobias Schnabel, Igor Labutov, David Mimno, and Thorsten Joachims.
\newblock 2015.
\newblock Evaluation methods for unsupervised word embeddings.
\newblock In \emph{Proceedings of the Conference on Empirical Methods in
  Natural Language Processing (EMNLP)}.

\bibitem[Socher et~al.(2011)Socher, Pennington, Huang, Ng, and
  Manning]{socher_et_al_2011}
Richard Socher, Jeffrey Pennington, Eric~H. Huang, Andrew~Y. Ng, and
  Christopher~D. Manning.
\newblock 2011.
\newblock Semi-supervised recursive autoencoders for predicting sentiment
  distributions.
\newblock In \emph{Proceedings of the Conference on Empirical Methods in
  Natural Language Processing}, EMNLP '11. Association for Computational
  Linguistics.

\bibitem[Tellex et~al.(2003)Tellex, Katz, Lin, Fernandes, and
  Marton]{tellex_et_al_2003}
Stefanie Tellex, Boris Katz, Jimmy Lin, Aaron Fernandes, and Gregory Marton.
\newblock 2003.
\newblock Quantitative evaluation of passage retrieval algorithms for question
  answering.
\newblock In \emph{Proceedings of the 26th Annual International ACM SIGIR
  Conference on Research and Development in Informaion Retrieval}, SIGIR '03.
  ACM.

\bibitem[Turian et~al.(2010)Turian, Ratinov, and Bengio]{turian_et_al_2010}
Joseph Turian, Lev Ratinov, and Yoshua Bengio.
\newblock 2010.
\newblock Word representations: A simple and general method for semi-supervised
  learning.
\newblock In \emph{Proceedings of the 48th Annual Meeting of the Association
  for Computational Linguistics}, ACL '10. Association for Computational
  Linguistics.

\bibitem[Turney and Pantel(2010)]{turney_and_pantel_2010}
Peter~D. Turney and Patrick Pantel, 2010.
\newblock From frequency to meaning: Vector space models of semantics.
\newblock Available at \url{http://arxiv.org/abs/1003.1141}.

\end{thebibliography}


\end{document}